\documentclass[a4paper,runningheads]{llncs}
\usepackage[T1]{fontenc}
\usepackage{graphicx}
\usepackage{caption}
\usepackage{subcaption}
\usepackage{hyperref}
\usepackage[inline]{enumitem}
\usepackage{algorithm}
\usepackage{algpseudocode}

\begin{document}
\title{The RoSiD Tool: Empowering Users \\ to Design Multimodal Signals for Human-Robot Collaboration}
%
%
\author{Nathaniel Dennler\inst{1} \and David Delgado\inst{1}  \and Daniel Zeng\inst{1}  \and \\
Stefanos Nikolaidis\inst{1} \and
Maja Matari\'c\inst{1}}
\authorrunning{N. Dennler et al.}
%
\institute{University of Southern California, Los Angeles CA
\email{\{dennler,nikolaid,mataric\}@usc.edu}\\
\url{http://www.springer.com/gp/computer-science/lncs}
}
\maketitle              
\begin{abstract}
Robots that cooperate with humans must be effective at communicating with them. However, people have varied preferences for communication based on many contextual factors, such as culture, environment, and past experience. To communicate effectively, robots must take those factors into consideration. In this work, we present the Robot Signal Design (RoSiD) tool to empower people to easily self-specify communicative preferences for collaborative robots. We show through a participatory design study that the RoSiD tool enables users to create signals that align with their communicative preferences, and we illuminate how this tool can be further improved.
\vspace{-.2cm}
\keywords{Human-robot Interaction  \and Personalization \and Signalling.}
\vspace{-.2cm}
\end{abstract}

\section{Introduction}

For robots to be effective collaborative partners, they must communicate information about their current state, task completion, and knowledge. People are usually effective at using different signals during collaboration \cite{knoblich2011psychological}; however, designing signals that allow robots to be effective remains challenging. People excel at adapting the way they communicate to their environment and collaborators \cite{knoblich2011psychological}. For robots to be effective, they must similarly adapt to a variety of contextual factors, however, robots have the additional challenge of understanding how their own embodiment affects how people expect to interact with them \cite{dennler2023design}. To address this problem, we aim to develop a way for people to encode the important contextual factors by allowing them to rapidly design signals themselves.

This work introduces the Robotic Signal Designer (RoSiD) tool we developed based on insights from exploratory research in human-computer interaction (HCI) and preference learning research in human-robot interaction (HRI). RoSiD facilitates the design embodied signals with three kinds of signal components used in robotic systems: visual, auditory, and kinetic. Importantly, RoSiD is an HRI tool that involves communication channels beyond HCI due to the robot's embodiment and ability to interact with the user and objects in the physical world, resulting in well-documented improvements in user engagement and task performance \cite{deng2019embodiment}. 

We explored three user study hypotheses related to system use characteristics to evaluate RoSiD:

\begin{enumerate}[label={\textbf{H\arabic*:}}]
    \item Participants will rate the system as usable according to the System Usability Scale \cite{brooke1996sus}.
    \item Participants will spend the most time designing the first signal to learn how to use the system.
    \item Participants will benefit more from having suggested signals based on the signals other participants designed than random signals.
\end{enumerate}


\section{Designing RoSiD}

\begin{figure}[t]
    \centering
     \begin{subfigure}[b]{0.37\textwidth}
         \centering
         \includegraphics[width=\textwidth]{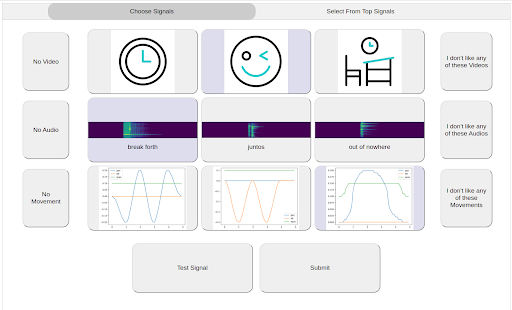}
         \caption{Query-based interface for choosing among three signals per modality.}
         \label{fig:select-page}
     \end{subfigure}
     \hfill
     \begin{subfigure}[b]{0.37\textwidth}
         \centering
         \includegraphics[width=\textwidth]{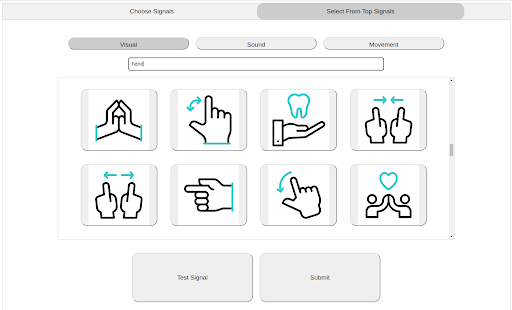}
         \caption{Search-based interface for browsing all options for each modality.}
         \label{fig:browse-page}
     \end{subfigure}
     \hfill
     \begin{subfigure}[b]{0.23\textwidth}
         \centering
         \includegraphics[width=\textwidth]{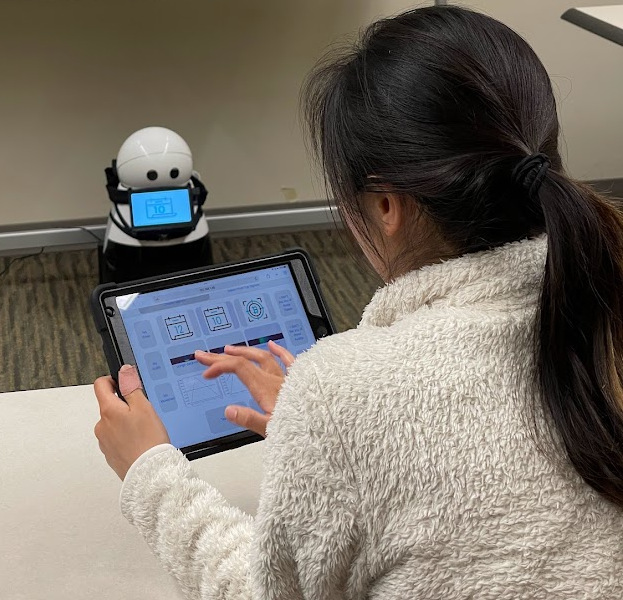}
         \caption{Participant using the RoSiD tool in our study.}
         \label{fig:setup}
     \end{subfigure}
     \hfill
      \caption{Interfaces for the RoSiD tool.}
\end{figure}

Following previous work, we considered robot signals as multimodal behaviors that consist of visual, auditory, and kinetic components. For each type of stimulus, we collected a large dataset of viable options from public websites. 
Specifically, we used 5,912 animated videos that represented the visual components, 867 sound clips that represented auditory components, and 2,125 head motions that represented the kinetic components\footnote{All files are publicly available on \href {https://github.com/ndennler/kuri_personalization/tree/main/stimuli}{github}.}. Based on the literature in preference learning and exploratory search, \cite{allen2021engage,brown2020safe,chang2019searchlens,sadigh2017active}, we employed two main interactions to select from these options: query-based and search-based interactions.

{\it Query-based interactions} are often used to learn user preferences in the field of human-robot interaction \cite{sadigh2017active}. In these interactions, users review a small number of robot behaviors and specify the behavior they think is best-suited for a given task. The behavior that the user selects from the small set of behaviors provides information about what they would like the robot to do in its particular context. The formulation of how preferences are modeled is provided in Section \ref{sec:user_prefs}. Our query-based interface is shown in Figure \ref{fig:select-page}. In this work, a single query, $Q$, consists of three specific videos played on Kuri's screen, sound clips played through Kuri's speakers, or motions played on Kuri's head. We include an option for the user to specify that none of the three items in the query are what they are looking for.

{\it Search-based interactions} are used in exploratory search contexts in human-computer interaction\cite{chang2019searchlens}. In these interactions, users are presented with a large number of possible options that can be filtered with key words. The order the options are presented in is important \cite{allen2021engage}. We use the preference data from the query-based interaction to inform the order of the search-based results. Our search-based interface is shown in Figure \ref{fig:browse-page}.

\section{Technical Approach}

\subsection{Understanding User Preferences}\label{sec:user_prefs}
We adopt the formulation of {\it preference learning}, where preferences are represented as a linear combination of a set of features that describe a time-series, as described by Sadigh et al. \cite{sadigh2017active}. Our goal is to learn the parameterization of the user's preference, $\omega$. We evaluate how well a particular query aligns with a user's preferences (to assess \textbf{H3}) using the following alignment metric inspired by \cite{sadigh2017active}:

\begin{equation}\label{equation:alignment}
    alignment = \mathbf{E} \left[ \max\limits_{q \in Q} \left(   \frac{\phi_q \cdot \phi_{selected}}{|\phi_q| \cdot |\phi_{selected}|} \right) \right]
\end{equation} 

where q represents the element in the query $Q$ (consisting of 3 items per modality in our system), and $\phi$ denotes the features of the particular stimulus, with $\phi_{selected}$ representing the features of the stimulus the participant selected at the end of the experiment. The maximum alignment score is 1 and the minimum alignment score is -1.

\subsection{Creating Features for Multimodal Data}
Our assessment relies on the stimuli used in our experiments being represented by a vector that encapsulates the characteristics of the stimulus (i.e., $\phi$ in Equation \ref{equation:alignment}). We chose to use a learned encoding from pretrained models, as non-linear features have been shown to be effective for preference learning \cite{brown2020safe}. All embeddings were reduced to 32 dimensions using PCA because dimension largely affects speed in preference learning, and the system was designed to run in real time.

\textit{Visual:} To create embeddings for the visual features, we used embeddings from a pretrained CLIP model available from the transformers library \cite{wolf2019huggingface}. Each video had a representative frame selected as the image component, and a short description used as a language component. 

\textit{Auditory:} Embeddings for the auditory features were generated by encoding our audio files with the pretrained VGGish model \cite{hershey2017cnn}. 

\textit{Kinetic:} Embeddings for kinetic features came from a GRU model trained through a Seq2Seq task \cite{sutskever2014sequence} on our movement data, where the series of states of the robot's head  (pan, tilt, eyes) were encoded through a recurrent network, and a second recurrent network was initialized with the embedding to reproduce the original sequence. 

\subsection{Generating Queries from User Data}\label{clustered_queries}
To address \textbf{H3}, we propose a method to generate queries $Q$ for signal design that contain items that are more aligned with what the users ultimately choose. For each signal, we have a dataset for each modality $\mathcal{D}$ that contains the final items selected by the users.

We base this approach on the insight that user preferences are a smaller set of all possible items in our datasets of signal components. We attempt to find clusters in preferences from the signals that users designed by using RoSiD. To do this, we partition $\mathcal{D}$ into k groups based on the features of the signal components, $\phi$. We then randomly select an item from each of these queries to create more meaningful suggestions. This process is outlined in Algorithm \ref{alg:query_gen}.

\begin{algorithm}
\caption{Generating queries from user data}\label{alg:query_gen}
\begin{algorithmic}[1]
\State \textbf{Input:} $\mathcal{D}$, a dataset of designed signals; $k$, the number of items in the resultant query; $cluster(\mathcal{D},k)$, a partitioning method that returns $f:\mathcal{D}\rightarrow\{1,2,...,k\}$;

\State \textbf{Output:} $Q$, a set of $k$ options for the user to select from when designing signals;

\State $Q \leftarrow \emptyset$; $f \leftarrow cluster(\mathcal{D}, k)$;

\For{$i \in \{1,2,...,k\}$}
\State $q_i \in_R \{d$  |  $\exists d \in \mathcal{D}, f(d) = i \}$
\State $Q \leftarrow Q \cup q_i$
\EndFor
\State \textbf{Return} $Q$

\end{algorithmic}
\end{algorithm}

\section{Design Session}
In this section, we describe the details of the interactions users had with the robot while using the RoSiD tool to design the robot's signals and evaluate those signals. Our protocols were approved by the university's Institutional Review Board under \#UP-23-00408.

\subsection{Study Description}\label{sec:study_description}
Participants engaged in a one-hour design session. Upon entering the experiment space, they were told that they would be designing four signals for a robot that will assist them with finding items around the experiment space. The signals consisted of three components: visual, auditory, and kinetic. 

\begin{figure}
    \centering
     \includegraphics[width=\linewidth]{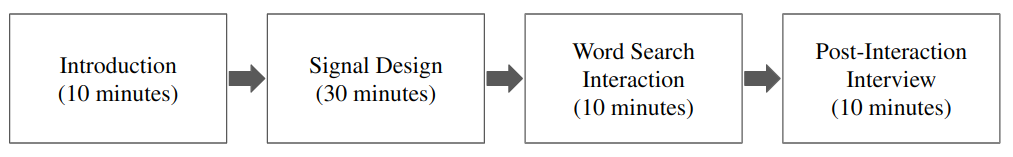}
     \caption{Structure of the design session with approximate times for each section.} \label{fig:study-flow}
\end{figure}

Participants designed signals for a modified Mayfield Kuri robot (shown in Figure \ref{fig:setup}). Since the robot does not have a screen or affordances for carrying items, we added an external screen to provide a salient visual component to the signalling and a backpack to hold the Raspberry Pi and power supply, with a pouch for holding objects being transported. The four signals participants designed were:

\begin{enumerate}
    \item \textit{Idle}: played every 10 seconds while the robot waits for commands, indicating that the robot is ready to accept a command.
    \item \textit{Searching}: played every 10 seconds while the robot searches for objects, indicating that the robot is actively searching for an item.
    \item \textit{Has Item}: played once, when the robot has an item in its pouch and is ready for the participant to remove the item.
    \item \textit{Has Information}: played once, when the robot has found an object, but the object is inaccessible. The participant can follow the robot to the location of the object to retrieve it.
\end{enumerate}

Each participants was then introduced to the RoSiD interface as described in Section \ref{sec:study_description} and designed the four signals in a randomized order to mitigate any ordering effects. The participant was free to use the interface however they liked, for as long as they liked. Participants tended to favor either the query-based and search-based interactions in their design process, but this was dependent on the individual. After finishing designing all four signals, the participant filled out the System Usability Scale \cite{brooke1996sus}. 

The participant next engaged in an interaction with the robot, where the robot was piloted by an experimenter. To simulate being occupied as the robot roamed around the environment, the participant was also engaged in a word search task. To complete the word search, the participant had to ask the robot to help them search for items around the room, which had words for the word search printed on the item. For example, participants were tasked to ask Kuri to find a stapler, and the stapler had the word "haptic" printed on it. The participant then located "haptic" in the word search. The time limit for this section was 10 minutes. Following the interaction, participants engaged in a semi-structured interview and were compensated with a 20 USD Amazon gift card. The entire study design is illustrated in Figure \ref{fig:study-flow}.

\subsection{Participants}

Participants were recruited from the USC student population through email, flyers, and word-of-mouth. A total of 25 participant were part of the study, with ages that ranged from 19 to 43 (median 25); participants self-declared as men (13), women (10), and genderqueer, nonbinary, or declined to state (3, aggregated for privacy; some participants belonged to multiple groups), 13 participants identified as LGBTQ+. All participants were able to create signals they liked for all four categories, and all successfully interacted with the robot to collect all the items in the word search task.

\section{Results}

\subsection{System Usability Scores}
We examined the participants' SUS scores based on recommendations from a meta-analysis of several extant systems \cite{lewis2018system}. The participants rated the system with a median score of 75 out of 100 on the SUS scale, suggesting that the system is between "good" and "excellent" on an adjective rating scale, and a letter grade of 'B' demonstrating an above-average user experience. Using a Mann-Whitney U-Test, we determined that the ratings were significantly higher than a 65 of 100 on the SUS scale (U = 10.0, $p=.015$), indicating that our system is above average in its ease of use, supporting \textbf{H1}.

\begin{figure}
    \centering
     \begin{subfigure}[b]{0.48\textwidth}
         \centering
         \includegraphics[width=\textwidth]{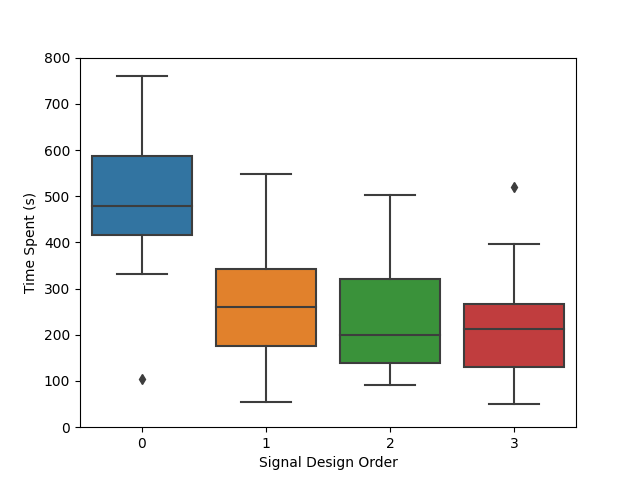}
         \caption{Time to design by order.}
         \label{fig:time-by-order}
     \end{subfigure}
     \begin{subfigure}[b]{0.48\textwidth}
         \centering
         \includegraphics[width=\textwidth]{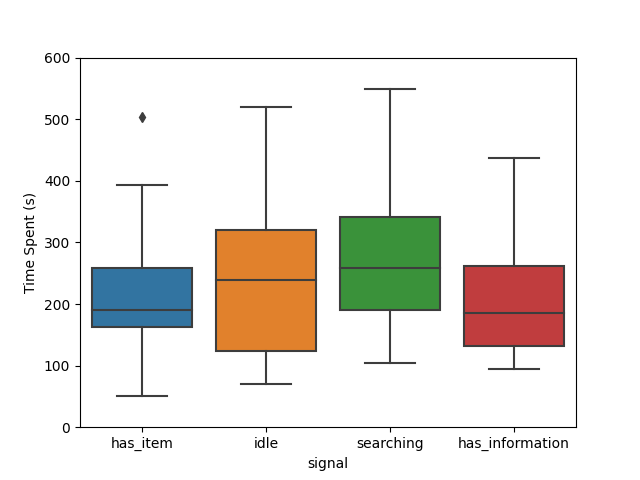}
         \caption{Time to design by signal.}
         \label{fig:time-by-signal}
     \end{subfigure}
      \caption{Box plots showing the times users spent deigning signals.}
\end{figure}

\subsection{Time Spent Designing Signals}
We examined how long it took users in our study to design the signals. An ANOVA revealed that the time to design signals depended on the order that they were designed in ($F(3,96)=26.549,p<.001$), as illustrated in Figure \ref{fig:time-by-order}. Post hoc analysis showed that the only significant pairwise differences were between the first signal designed and the rest. This indicates that our system is easy to learn to use, because the time to design signals stabilized after the first designed signal, supporting \textbf{H2}. We also found no significant differences between the kind of signal and the time to design the signal, illustrated in Figure \ref{fig:time-by-signal}, indicating that the signals were similarly easy to design. This implies that the particular signals we selected were easily understandable for the participants. The type of signal had little effect on the results of our analysis.

\begin{figure}
    \centering
     \begin{subfigure}[b]{0.32\textwidth}
         \centering
         \includegraphics[width=\textwidth]{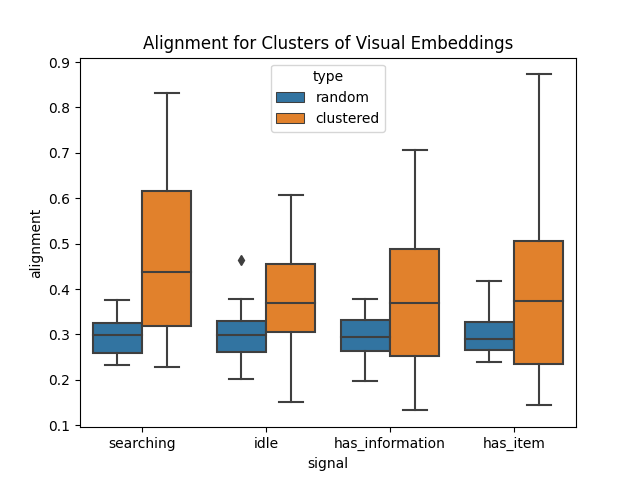}
         \caption{Visual components.}
         \label{fig:visual-improvement}
     \end{subfigure}
     \hfill
     \begin{subfigure}[b]{0.32\textwidth}
         \centering
         \includegraphics[width=\textwidth]{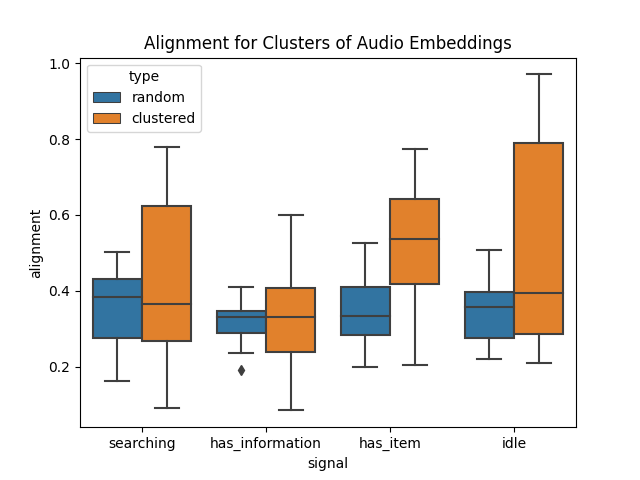}
         \caption{Auditory components.}
         \label{fig:auditory-improvement}
     \end{subfigure}
     \hfill
     \begin{subfigure}[b]{0.32\textwidth}
         \centering
         \includegraphics[width=\textwidth]{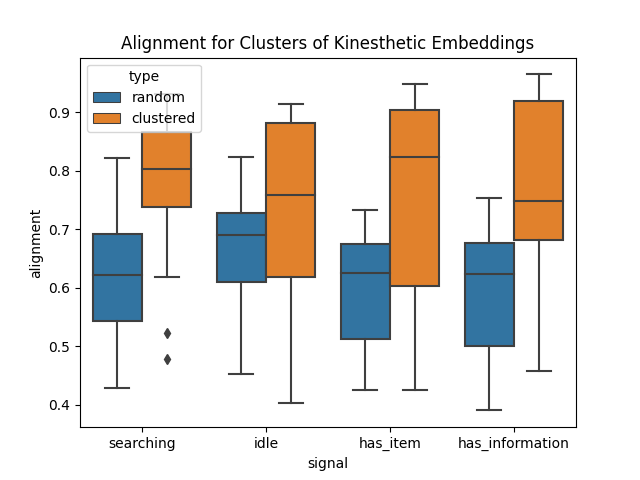}
         \caption{Kinetic components.}
         \label{fig:kinetic-improvement}
     \end{subfigure}
     \hfill
     \caption{Box plots comparing the alignment of initial queries based on random suggestions and the proposed clustered suggestions.}
\end{figure}

\subsection{Using Clusters to Initialize Queries}
We examined how we could use prior information based on the signals collected from other users to generate queries that are more aligned with what participants ultimately chose when designing their own signals. We used a leave-one-out cross-validation setting for each participant and formed clusters from all but one participant following the process in Section \ref{clustered_queries}. For our clustering method we used agglomerative clustering as implemented in scikit-learn \cite{scikit-learn}. We calculated the alignment score as described in Section \ref{sec:user_prefs} for the clustering method as compared to randomly selecting queries for each of the three modalities. We performed an ANOVA analysis for each of the modalities to study the effect of including other user's information on the maximum query alignment for new users.

We found that for the visual modality there is a significant main effect across query method ($F(1,3)=44.106,p<.001$), with an average increase in initial alignment of .117 across all signals when using the clustering method over randomly selecting stimuli. For the auditory modality there was a significant main effect of query method ($F(1,3)=19.544,p<.001$), with an average increase in initial alignment of .141 across all signals. For the kinetic modality there was also a significant effect of query type ($F(1,3)=49.393,p<.001$). For the kinetic modality there was an average increase in initial alignment of .132 across all signal types.

\section{Conclusions and Future Work}
In this work we developed the RoSiD tool, which enables users to design their own robot signals for collaborative tasks with robots. Our results show that users find this system easy to use, quick to learn, and that using past user data can further improve the system's usefulness. In continued work, we plan to further develop the RoSiD tool and its potential for use with other robot embodiments using the insights gained from this study. We will evaluate the improved tool by measuring performance and other behavioral metrics in a real item-finding task, as well as compare the effect of using personalized signals in contrast to using generic signalling methods.

%

\bibliographystyle{splncs04}
\bibliography{mybibliography}

\end{document}